\documentclass[letterpaper, 10 pt, conference]{ieeeconf}  

\IEEEoverridecommandlockouts                              

\usepackage{graphicx} 
\usepackage{amsmath} 
\usepackage{amssymb}  
\usepackage[algo2e, ruled,vlined]{algorithm2e}
\usepackage{algorithm}
\usepackage[noend]{algpseudocode}
\usepackage{siunitx}
\usepackage{hyperref}
\usepackage{bm}
\usepackage{xcolor}

\title{\LARGE \bf
Learning Agile Locomotion Skills with a Mentor}

\author{Atil Iscen$^{1*}$, George Yu$^{1*}$, Alejandro Escontrela$^{1,2}$, Deepali Jain$^{1}$, Jie Tan$^{1}$ and Ken Caluwaerts$^{1}$
\thanks{$^{*}$Equal contribution \& corresponding authors. }
\thanks{$^{1}$ Robotics at Google.
        {\tt\small \{atil,georgeyu,aescontrela, jaindeepali,jietan,kencaluwaerts\}@google.com}}%
\thanks{$^{2}$ Georgia Institute of Technology. 
        {\tt\small aescontrela@gatech.edu}}%
}

\begin{document}

\maketitle
\thispagestyle{empty}
\pagestyle{empty}

\begin{abstract}
Developing agile behaviors for legged robots remains a challenging problem. 
While deep reinforcement learning is a promising approach, learning truly agile behaviors typically requires tedious reward shaping and careful curriculum design. 
We formulate agile locomotion as a multi-stage learning problem in which a mentor guides the agent throughout the training. 
The mentor is optimized to place a checkpoint to guide the movement of the robot's center of mass while the student (i.e. the robot) learns to reach these checkpoints. 
Once the student can solve the task, we teach the student to perform the task without the mentor. We evaluate our proposed learning system with a simulated quadruped robot on a course consisting of randomly generated gaps and hurdles. Our method significantly outperforms a single-stage RL baseline without a mentor, and the quadruped robot can agilely run and jump across gaps and obstacles. Finally, we present a detailed analysis of the learned behaviors' feasibility and efficiency.
\end{abstract}

\section{Introduction}
Developing agile behaviors for legged robots remains a challenging problem. Agile locomotion skills require fast reactions, coordinated control of legs, precise manipulation of contact forces, and robust balance control. Hand-engineering such controllers requires significant expertise and often tedious manual tuning. Deep Reinforcement Learning (DRL) is a promising approach that can acquire locomotion controllers automatically. However, learning truly agile motions over complex terrains, such as jumping over gaps, remains an open problem.  

This paper formulates the challenge of learning agile locomotion with a mentor-student relationship and uses a multi-stage reinforcement learning approach to tackle the problem. In this process, a mentor learns to perform a task and reward shaping by placing \emph{checkpoints} to guide the movement of the robot's center of mass, while the student, the robot itself, learns to complete the task by collecting these checkpoints. 

We divide the training pipeline into three stages. In the first stage, a simplified task is presented to train the student and find the task's best mentor. In the second stage, we focus on generalization: we introduce randomized and more challenging tasks and deploy a curriculum to overcome the increased difficulty. In the last stage, the robot learns to rely on its perception input to traverse the terrain, with decreasing, and eventually zero reliance on the mentor. After the last stage, the robot can traverse the terrain agilely without using privileged information provided by the mentor. Each stage of training is warm-started using the best policy from the previous stage.

\begin{figure}[t]
    \centering
    \includegraphics[width=1.0\linewidth]{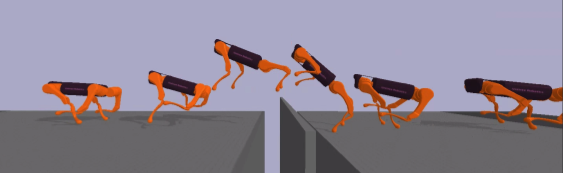}
    \caption{Snapshot of the agile and natural jumping behaviors that emerge from policies trained using our three-stage method. }
    \label{fig:teaser}
\end{figure}

Figure \ref{fig:teaser} shows the resulting behaviors from our trained policies. We show that our method significantly outperforms a single-stage RL baseline without a mentor, which is stuck at a local maxima caused by the gap. The main contributions of our paper include:

\begin{enumerate}
    \item A novel formulation of developing agile locomotion controllers as a reinforcement learning problem with a mentor-student relationship, which significantly improves the success rate of learning;
    \item A three-stage training pipeline in which agile locomotion skills emerge automatically;
    \item A detailed analysis of a quadruped robot's learned motion across a challenging obstacle course in simulation.
\end{enumerate}

\section{Related Work}
In decades of research on legged robots~\cite{raibert1986legged}, researchers have explored different gaits~\cite{park2017high,hyun2016implementation} and morphologies~\cite{briggs2012tails, patel2014rapid, khoramshahi2013piecewise, eckert2015comparing} to produce agile behaviors that mimic those observed in the animal kingdom. Using simplified dynamics and model-predictive control (MPC), the MIT Cheetah can run at $6\si{\metre\per\second}$ and jump over $0.4\si{\metre}$ obstacles~\cite{park2015online, nguyen2019optimized}. Similar techniques have enabled ATLAS~\cite{bd2018}, a humanoid robot, to backflip and even parkour across an obstacle course. These optimal control-based techniques often require an accurate model, deep prior knowledge, and tedious tuning. To reduce the amount of manual work, heuristics were extracted to regularize the MPC formulation, which enabled agile gaits on the MIT Mini Cheetah platform~\cite{bledt2020extracting}. Lately, MPC has also been combined with learned models to achieve faster on-robot learning of locomotion~\cite{yang2019data}.

Researchers have recently successfully applied deep reinforcement learning to automatically learn locomotion controllers~\cite{tan2018sim2real, hwangbo2019learning, haarnoja2018learning, xie2019iterative, RoboImitationPeng20, Lee_2020}. Numerous agile behaviors were demonstrated in simulation~\cite{2016-TOG-deepRL, peng2015dynamic, heess2017emergence, xie2020allsteps}. However, learning agility, even in simulation, still presents many challenges. Extensive reward shaping, expert demonstrations~\cite{RoboImitationPeng20,2018-TOG-deepMimic}, curriculum learning~\cite{xie2020allsteps}, and multi-agent learning~\cite{tang2020learning} are often needed for successful training. Tang et al.~\cite{tang2020learning} formulate learning agility as an adversarial multi-agent game, inspired by the pursuit-escape behaviors in nature to encourage the emergence of agile running gaits. The concept of mentor-student in our method resembles a multi-agent learning setting. A key difference in our formulation is that \emph{the mentor and the student work collaboratively to accomplish a common goal: the student successfully acquires agile gaits}. This makes our method simpler and more robust because training an ensemble of adversaries to stabilize learning is unnecessary in a collaborative learning environment. As a result, our method can tackle more complex tasks. Our idea that the mentor provides the student with privileged information to help it learn is similar to ``Learning by Cheating''~\cite{chen2019learning}. However, we use a different approach to remove the need for this privileged information at the last stage of our pipeline.

\section{Problem Definition}

Our goal is to train an agile locomotion policy with which a quadruped robot can jump over randomly placed gaps and hurdles. 
As in standard RL environments formulated as Markov Decision Processes, we define the problem in terms of state, reward and actions. At any time step $t$, the robot takes an action $\bm{a}_t \in A$ and it receives a set of observations ($\bm{s}_t \in S$) and an associated reward ($r_t \in \mathbb{R}$).  The state is composed of
$$\bm{s}_t = [\bm{g}^T_t,\bm{o}_t^T,\bm{a}_{t-1}^T]^T,$$
where $\bm{o}_t = [\bm{o}_{{\tiny {\mbox{ENC}}},t}^T, \bm{o}_{{\tiny {\mbox{IMU}}},t}^T, \bm{o}_{{\tiny \mbox{lidar}},t}^T]^T$ contains the sensor observations (motor encoders, IMU and LiDAR, see Section~\ref{Sec:exp_setup}). $\bm{g}_t = [g_{d,t}, g_{h,t}, g_{z, t}]^T$ includes the distance $g_d$, relative heading $g_h$, and relative height $g_z$, to the target position, and $\bm{a}_{t-1}$ is the previous action taken by the robot.

A basic locomotion task is to achieve forward motion while maintaining the balance of the robot. More complex tasks consist of uneven terrains or obstacles that require more advanced maneuvers or skills. In this work, we focus on the agility aspect of quadruped locomotion, where the desired tasks require speed, fast response, and a combination of different types of behaviors based on the perception of the environment. These agile behaviors require gaits with long flight phases instead of (quasi-)static gaits where multiple feet stay in contact with the ground at any given time.

To train and test our algorithm, we designed a course segment $2\si\metre$ in width and $6\si\metre$ in length in which the robot moves from one end to the other. The course contains a large gap ($0.9\si{\metre}$) that the robot has to cross to reach the end. The desired final behavior consists of running at high speed, transitioning to jumping based on sensed distance to the gap, a flight phase, landing, and transitioning back to running. In another more complex scenario, we also introduce a vertical hurdle that requires more vertical motion than the forward leap. Getting across the hurdle requires accurately timing the jump at the right place and with the right pose to pass the hurdle. The reward function consists of two components that provide a dense and a large sparse reward. The dense reward is the negative change in distance to the goal location, and the sparse reward is a large bonus for reaching the target location:
$$
r_t = (g_{d,t} - g_{d,t-1}) + B_g I(g_{d,t}\leq r_g),
$$
where $g_{d,t}$ is the distance to the goal, $B_g$ is the bonus for reaching the target, $r_g$ is the threshold distance for reaching the target, and $I(\cdot)$ is the indicator function.

The proposed two tasks (gap and gap with a hurdle) pose a  challenge to RL methods due to the existence of a steep local maximum (being unable to cross the gap), lack of ground truth observations (e.g. heightmap), multi-modal observations (LiDAR, IMU and motor encoders) and the need for a transition between diverse behaviors (i.e. running, jumping, landing). To tackle this challenge, we apply a three-staged training approach, detailed in the Methods section below.

\section{Method}
To handle the challenges posed by the agile locomotion objective, we propose a method that uses a combination of:
\begin{itemize}
    \item Multi-stage population-based training.
    \item Task and reward shaping using a mentor.
    \item Curriculum to master hard tasks and to improve generalization.
\end{itemize}

We first approach the problem by sequencing the training into different stages, to focus on solving each subproblem, and build up to the final task. In particular, we propose a three-staged learning framework to train the policy, where each stage narrows the problem scope. Our method is summarized in Algorithm~\ref{alg}. We define the three training stages as:

\begin{enumerate}
    \item Defining the right mentor on a simplified problem.
    \item Generalization to the original problem scope.
    \item Removing the dependence on the mentor.
\end{enumerate}
Between individual stages, we carry only the best policy of the population to the next stage, and warm start the next stage of training. 

\begin{figure}[t]
    \centering
    \includegraphics[width=0.6\linewidth]{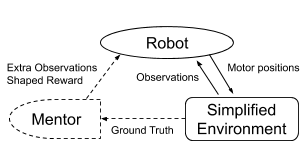}
    \caption{Our method introduces a mentor to the system in stage 1 of the training. The mentor provides reward shaping and extra observations based on the ground truth. Later in the training (stage 3), the robot learns not to rely on these observations and we remove the mentor from the system.}
    \label{fig:mentor}
\end{figure}
\begin{figure}[t]
    \centering
    \includegraphics[width=0.9\linewidth]{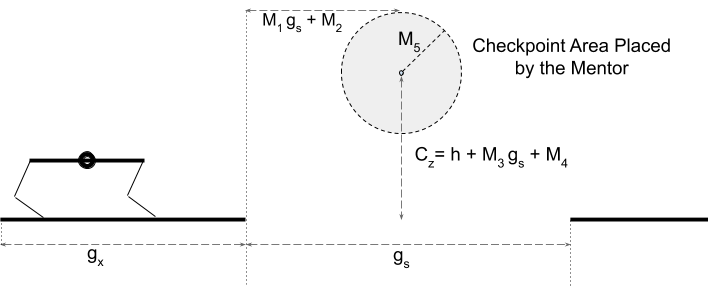}
    \caption{Illustration of the checkpoint area defined by the mentor's parameters. Checkpoint area's position and radius are defined using 5 parameters ($\bm{m}=M_1$..$M_5$). These parameters are optimized as a hyperparameter in the first stage of the training. The checkpoint is then removed using dropouts during the third stage of the training.}
    \label{fig:gap_mentor_params}
\end{figure}

In the first stage, we simplify the problem by fixing the location of the gap with respect to the robot's starting position. Specifically, the start of the gap is positioned $2\si{\metre}$ ahead of the robot. Learning starts with a small ($0.3\si{\metre}$) gap. As training progresses, the size is randomly sampled from a distribution where we gradually step up the difficulty by linearly increasing the upper bound to $0.9\si{\metre}$ based on the number of training steps. This process is referred to as \emph{curriculum(lower bound, upper bound)} in the pseudocode. In addition to simplification of the problem, the main goal of the first stage is to introduce the concept of the mentor to the robot. The mentor and robot collaborate to achieve maximal reward on the task. In particular, the mentor aids the robot by providing (1) both reward and task shaping in form of a checkpoint  (2) additional privileged information in the form of ground-truth observations. 

The mentor is responsible for selecting an intermediate checkpoint location that the robot has to reach. The robot collects dense rewards for getting closer to and a large reward when it reaches the checkpoint. One could allow the mentor to adapt online, making decisions at each time step, which would make the problem a multi-agent system with simultaneous learners. Instead, for simplicity, we set the mentor to be constant and define it as a set of hyperparameters ($\bm{m}=\{M_i\}_{i=1,...,5} \in \mathbb{R}^5$) for the RL algorithm. At every time step, the mentor provides the checkpoint's position in space relative to the robot's position and orientation. This additional information is appended to the robot's original observations (Fig. \ref{fig:mentor}). The mentor proposes the checkpoint location based on the gap location and the gap size (Fig. \ref{fig:gap_mentor_params}). Specifically,
$$ C_x = g_x + g_s  M_1 + M_2, $$
$$ C_z = h + g_s  M_3 + M_4,$$
$$ r_g = M_5. $$
$C_x$ and $C_z$ indicate the checkpoint's x and z coordinates, $g_x$ and $g_s$ stand for the gap's starting location (x coordinate) and size, $h$ is a constant for default walking height ($0.3\si{\metre}$). The last variable $M_5$ is the checkpoint radius, which is the threshold distance that determines whether the robot has successfully collected the checkpoint. Once the checkpoint is reached, the extra reward is collected and the checkpoint is assigned to the end of the track as goal location for the rest of the episode. The mentor and gap configurations are defined in the x-z plane. However, the robot is not constrained to this vertical plane and learning to robustly move forward is part of the task.

The second stage of the training is about generalization from the simpler task to the original task where the gap location is not fixed. This stage addresses the subproblem of stochasticity in the gap position--the problem is significantly harder, because the robot needs to adjust its running behavior prior to reaching the gap and avoid a  flight phase just before the gap. This stage can be considered as learning to transition to a jumping position at arbitrary gap placements. We warm start all the policies with the best policy of the previous stage and fix the mentor to the best set of hyperparameters from the first stage. The location of the gap is chosen randomly at each episode, and the range of the possible values is drawn from a uniform distribution $g_x \sim U(l,h)$, where the upper bounds $h$ is increased linearly  based on the training steps. At the end of the second stage, we obtain policies that can cross gaps positioned at different locations along the path.

\begin{algorithm} [t]
\mbox{\textbf{Stage 1:} Find the right mentor.}\\
$g_s \leftarrow$  curriculum($0.3\si{\metre}, 0.9\si{\metre}$)\\
$g_x \leftarrow$ $2.0\si{\metre}$\\
\For{$i \in [1..90]$}{
    $\bm{m}_i \leftarrow $ uniform(${M_{1..5}}$) // sample random mentor\\
    $R_{i}^{\mbox{{\tiny stage 1}}}, \pi_{i} \leftarrow $ train($\pi_{0},\bm{m}_i,g_s,g_x$) {\small //in parallel $\forall i$}
}
$\bm{m}_{{\scriptsize\mbox{best}}}, \pi_{{\scriptsize \mbox{best}}} \leftarrow$ $\mbox{argmax}_i(R_{i}^{\mbox{{\tiny stage 1}}}$)\\
\mbox{\textbf{Stage 2:} Generalization to original problem scope.}\\
$g_s \leftarrow$  $0.9\si{\metre}$\\
$g_x \leftarrow$ curriculum($2.0\si{\metre}, 3.0\si{\metre}$)\\
\For{$j \in [1..30]$}{
    $R_{j}^{\mbox{{\tiny stage 2}}}, \pi_j$ $\leftarrow$ train($\pi_{{\scriptsize\mbox{best}}}$,$\bm{m}_{{\scriptsize\mbox{best}}}$,$g_s$,$g_x$) {\small //in parallel $\forall j$}
}
$\bm{m}_{{\scriptsize\mbox{best}}}, \pi_{{\scriptsize \mbox{best}}} \leftarrow \mbox{argmax}_j(R_j^{\mbox{{\tiny stage 2}}})$\\
\mbox{\textbf{Stage 3:} Remove dependency on mentor.}\\
$P_d\leftarrow$  curriculum($0.0, 1.0$) // $\bm{g}_t$ dropout\\
$g_s \leftarrow$  $0.9\si{\metre}$\\
$g_x \sim U(2.0\si{\metre}, 3.0\si{\metre}$)\\
\For{$k \in [1..30]$}{
    $R_{k}^{\mbox{{\tiny stage 3}}}, \pi_k \leftarrow$ train($\pi_{{\scriptsize \mbox{best}}}$,$\bm{m}_{{\scriptsize\mbox{best}}}$,$g_s$,$g_x$) {\small //in parallel $\forall k$}
}
$\pi_{{\small \mbox{final}}} \leftarrow \mbox{argmax}_k(R_k^{\mbox{{\tiny stage 3}}})$\\
\caption{Training using a mentor.}
\label{alg}
\end{algorithm}

In the third stage, we aim to eliminate the student's dependency on the mentor. The student policy needs to rely solely on its proprioceptive sensors (LiDAR, IMU and motor encoders in our case) instead of the privileged checkpoint information fed by the mentor. We approach this problem as \emph{dropping out} the mentor. At each time step, the dropout probability ($P_d$) determines if the observations provided by the mentor should be replaced by dummy values. We use a curriculum to gradually increase $P_d$ as the number of training steps increases. Throughout the curriculum learning, the robot gradually learns to rely on its own sensors instead of the mentor's increasingly noisy guidance. At the end of the curriculum, the dropout is $100\%$ where the robot learns to complete the task without any help from the mentor. 

Every timestep, our controller obtains LiDAR scans ($\bm{o}_{{\tiny \mbox{lidar}},t}$), IMU readings ($\bm{o}_{{\tiny \mbox{IMU}},t}$) and joint angles ($\bm{o}_{{\tiny \mbox{enc}},t}$) while it outputs the desired motor positions ($\bm{a}_t$) which are converted to torques using a PD controller. Within the controller, we use a trajectory generator the same way as it was introduced by Iscen et al in \cite{iscen2018policies}. Policies Modulating Trajectory Generators (PMTG) is based on the cyclic motion of the legs using prior knowledge, but it still provides the freedom to achieve agile behaviors using TG's parameters and residual actions.

\begin{figure}[t]
    \centering
    \includegraphics[width=1.0\linewidth]{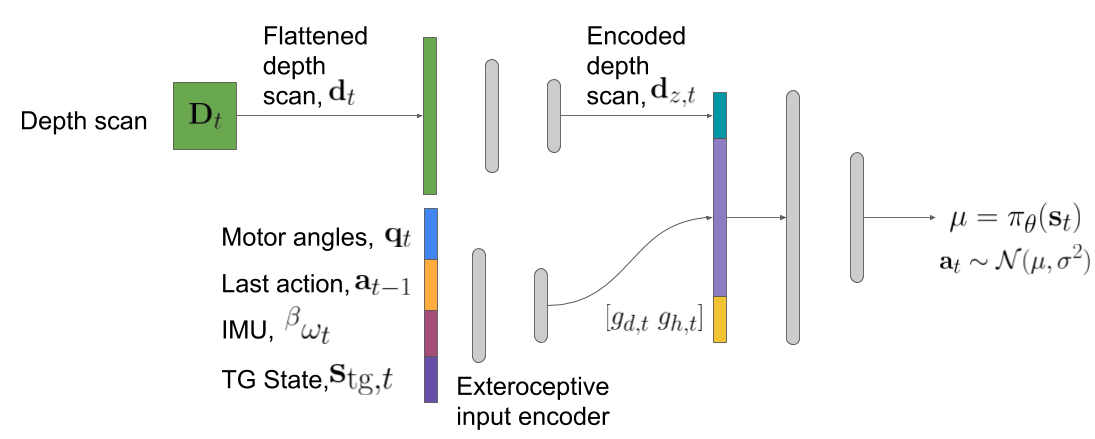}
    \caption{Policy Architecture. Depth scan provided by the LiDAR is flattened and separately encoded before being merged with other inputs. Value function and policy are parameterized by 2-layer networks of dimensions $(512, 256)$ and $(256, 128)$, respectively, with ReLU activations. LiDAR is fed into an encoder of dimensions $(32, 16, 4)$ and then passed to the policy.}
    \label{fig:architecture}
\end{figure}

At the heart of our controller, we have the neural network (NN) which is trained via RL or ES. We chose different NN architectures based on the type of the inputs and the algorithm used. As the training algorithm, we use Proximal Policy Optimization (PPO)~\cite{schulman2017ppo} but also test our method with
Evolutionary Strategies (specifically Augmented Random Search (ARS)~\cite{ars2018}). The architecture designed for PPO has different components  based on the type of inputs (Fig. \ref{fig:architecture}). The LiDAR input has its own separate encoder with a bottleneck, while proprioceptive inputs such as IMU, motor angles, last action and TG state go through another encoder. The outputs of these two encoders are concatenated and fed into the policy network, which outputs a multivariate Gaussian distribution of actions to sample from. As detailed in \cite{heess2017emergence}, this architecture achieves a separation of concerns between the basic locomotion skills and terrain perception and navigation. This new architecture enables the agent to adapt its smooth locomotion behaviors to the surrounding terrain.  In contrast with PPO, we use a much simpler NN for ARS, where all the inputs are concatenated and passed through a 2-layer fully connected NN, where each layer contains 32 units.

\section{Experimental Setup}
\label{Sec:exp_setup}
The simulated quadruped robot has three actuators per leg (12 total), corresponding to abduction, hip, and knee joints. 
We simulate a small quadruped, similar in size, actuator performance, and range of motion to the MIT Mini Cheetah~\cite{minicheetah} ($9\si{\kg}$) and Unitree A1\footnote{\href{https://www.unitree.com/products/a1/}{https://www.unitree.com/products/a1/}} ($12\si{\kg}$)  robots, which have both demonstrated highly agile skills.
We use the Unitree A1's URDF description\footnote{\href{https://github.com/unitreerobotics}{https://github.com/unitreerobotics}}, which is available in the PyBullet simulator~\cite{pybulletcoumans}.

As we focus on agile skills, it is important to use realistic dynamics and actuator models that are physically plausible.
To this end, we combine our PyBullet simulation environment~\cite{tan2018sim2real} with a more advanced actuator model based on the specifications of the T-Motor AK80-6, which is a commercially available actuator with detailed specifications and similar characteristics to the Mini Cheetah's and A1 robot's actuators. 
Our actuator simulation is based around a linear DC motor model. To reproduce the actuator's performance near its rated velocity, we linearly reduce its torque constant between the rated speed $38.2\si{\radian\per\second}$ and the zero torque speed $44.5\si{\radian\per\second}$ (estimated). This model accurately reproduces the torque/speed curve provided by a T-Motor.
The peak output torque, rated speed and peak mechanical power of the actuator are approximately $12\si{\newton\metre}$, $38.2\si{\radian\per\second}$ and $458\si{\watt}$. Hence, our simulation is conservative compared to the Mini Cheetah ($17\si{\newton\metre}$, $40\si{\radian\per\second}$, $680\si{\watt}$~\cite{minicheetah}) and A1 ($33.5\si{\newton\metre}$, $21\si{\radian\per\second}$, $703\si{\watt}$). 
This level of detailed simulation is to ensure that our simulation results can be reproduced in hardware. However, we make no claims about the direct sim-to-real transfer performance of our simulation model.

Our policies compute joint target positions ($\bm{a}_t$), which are converted to target joint torques by a PD controller running at $1\si{\kilo\hertz}$.
Rigid body dynamics and contacts are simulated at $1\si{\kilo\hertz}$ as well and the internal actuator dynamics are simulated at $10\si{\kilo\hertz}$.
In other words, the position and velocity (provided by PyBullet) and the desired torque (provided by the PD controller) are sent to the actuator model every $1\si{\ms}$. The actuator model then computes $10$ internal $100\si{\us}$ steps and provides the effective output torque of the actuator, which is then used by PyBullet to compute joint accelerations.
The simulation environment is configured to use an \emph{action repeat} of $10$ steps, which means that our policies compute a new action $\bm{a}_t$ and receive a state $\bm{s}_t$ every $10\si{\ms}$ ($100\si{\hertz}$).

\begin{figure}[t]
    \centering
    \includegraphics[width=0.7\linewidth]{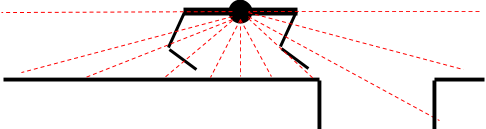}
    \caption{Simplified 2D illustration of the LiDAR observations $\bm{o}_{{\tiny \mbox{lidar}}}$.}
    \label{fig:lidar}
\end{figure}

As an exteroceptive sensor, we use a simulated LiDAR sensor to provide the agent with information of the surrounding terrain. This simulated LiDAR supports $20\times 20$ channels with a 360$^\circ$ horizontal and 180$^\circ$ vertical field of view (Fig.~\ref{fig:lidar}). We normalize the 3D depth scan to the range $[0, 1]$ and flatten to a vector ($\bm{o}_{{\tiny \mbox{lidar}}}$). 

In addition to the LiDAR, the robot has the following proprioceptive sensors:
\begin{itemize}
    \item IMU: roll, pitch, roll rate, pitch rate and yaw rate ($\bm{o}_{{\tiny \mbox{IMU}}} = [\phi,  \theta, \dot{\theta}, \dot{\phi}, \dot{\psi}]^T$).
    \item Motor angle encoders ($\bm{o}_{{\tiny \mbox{ENC}}}=[q_1, ..., q_{12}]^T$).
\end{itemize}


\section{Results}

\begin{figure}[tb]
    \centering
    \includegraphics[width=0.8\linewidth]{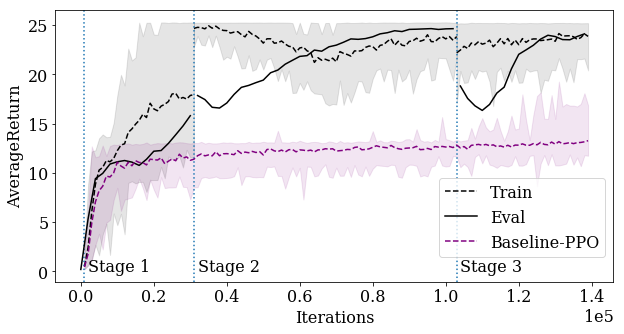}
    \caption{Learning curves of learning to cross a gap of 0.9 meters using PPO. We use three-stage training to find the mentor, generalization and mentor dropout while we warm start each stage with the best policy from the previous stage and use curriculum. 'Train' shows the return obtained during training, 'Eval' indicates evaluation over final state of the curriculum. The plots are averaged over 30 runs, shaded area indicates minimum and maximum of all the runs. Our framework learns to cross a gap and reach the top reward while the baseline PPO struggles at a local maxima.}
    \label{fig:learning-gap}
\end{figure}

\begin{figure}[t]
    \centering
    \includegraphics[width=0.8\linewidth]{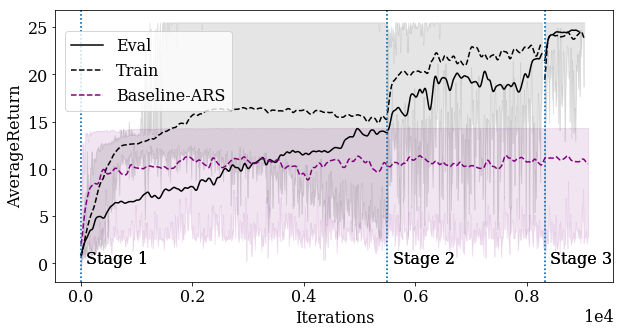}
    \caption{Learning curves for crossing a gap of $0.9\si{\metre}$ using ARS. We use three-stage training as in PPO. Each stage is warm started from the best policies from the previous stage. The dashed line is the training return, solid line is the evaluation, and shaded region represents the maximum and minimum rewards reached by the set of policies in each phase.}
    \label{fig:full_ars}
\end{figure}

\begin{figure}[t]
    \centering
    \includegraphics[width=0.8\linewidth]{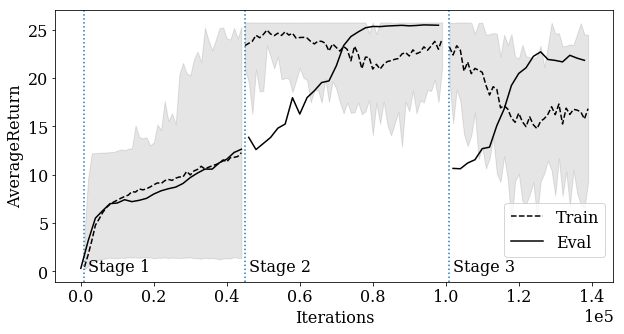}
    \caption{Learning curves for crossing a hurdle of 0.2m with a gap of 0.45m. The agent is trained the same way as Figure \ref{fig:learning-gap}, using PPO. The variance is larger due to increased complexity of the problem. Our framework reaches a high return accomplishing the task in most of the test cases.}
    \label{fig:learning-gap-hurdle}
\end{figure}

We design experiments to validate the efficacy of our training method, and also perform a detailed analysis of the learned agile gaits. In particular, we show that:

\begin{itemize}
    \item Our method significantly outperforms a single-stage RL baseline without a mentor;
    \item Our method is compatible with different learning algorithms;
    \item Our method leads the robot to learn various agile behaviors, based on its own perception;
    \item Our method results in policies with reasonable mechanical characteristics.
\end{itemize}


\subsection{Learning Results}
In Fig. \ref{fig:learning-gap}, we illustrate our results in learning to cross a gap with PPO, and show that our three-stage training method significantly outperforms a single-stage RL baseline without a mentor. Evaluations are performed against the end state of each stage.

\begin{itemize}
    \item Stage 1: Evaluate against gap at a fixed distance, with checkpoint observations.
    \item Stage 2: Evaluate against multiple different gap positions, with checkpoint observations.
    \item Stage 3: Evaluate against the final task: multiple different gap positions with no checkpoint observations.
    \item Baseline: Train without a mentor and evaluate against the final task: multiple different gap positions with no checkpoint observations.
\end{itemize}
The baseline gets stuck at a local maxima at the gap location, and fails to solve the task. In contrast, our method successfully accomplishes the task and achieves a high reward after the final stage.

The first stage learns to run and jump over a gap at a fixed location with the help of a mentor. We use 90 runs in parallel to test different hyperparameters as mentors. Due to differences of the mentors, the variation in the performance between the runs is rather significant in this stage. Some of the runs converge to achieve the task at approximately $30k$ simulation steps. We carry the best policy and the best mentor to the second stage, to learn to cross gaps that are randomly located. As the stochasticity grows, we see a temporary decrease in performance before a rebound when the policy adapts to the task. In the final stage, we gradually remove the mentor. After $40k$ more steps, the policies learn to achieve high returns without the help of the mentor. 

Next, we show that our method is not specific to a single learning algorithm. We also tackle the gap-jumping problem with ARS. Fig. \ref{fig:full_ars} presents the results using ARS on the same task as in Fig. \ref{fig:learning-gap}. Although in this experiment, we use a smaller 2-layer fully-connected NN as input to PMTG, we find that our three-stage training still yields significant improvements over the baseline. 



In Fig. \ref{fig:learning-gap-hurdle}, we demonstrate more impressive agile behaviors on a difficult extension of the gap jumping task. We fix the gap size to be $0.45\si{\metre}$, and place a hurdle of height $0.2\si{\metre}$ at the center of the gap. This task is more challenging as the robot cannot get to the other side with pure speed. The hurdle requires a significant ($0.3\si{\metre}$) vertical motion and a harder landing to accomplish the task. Our method successfully solves the task and achieves a high reward after the third stage of the training. Compared to Fig.~\ref{fig:learning-gap}, the variance of learning is larger, and returns drop lower at the start of second and third stages of the training, which is likely due to the increased difficulty of the task.


\begin{figure}[t]
    \centering
    \includegraphics[width=1.0\linewidth]{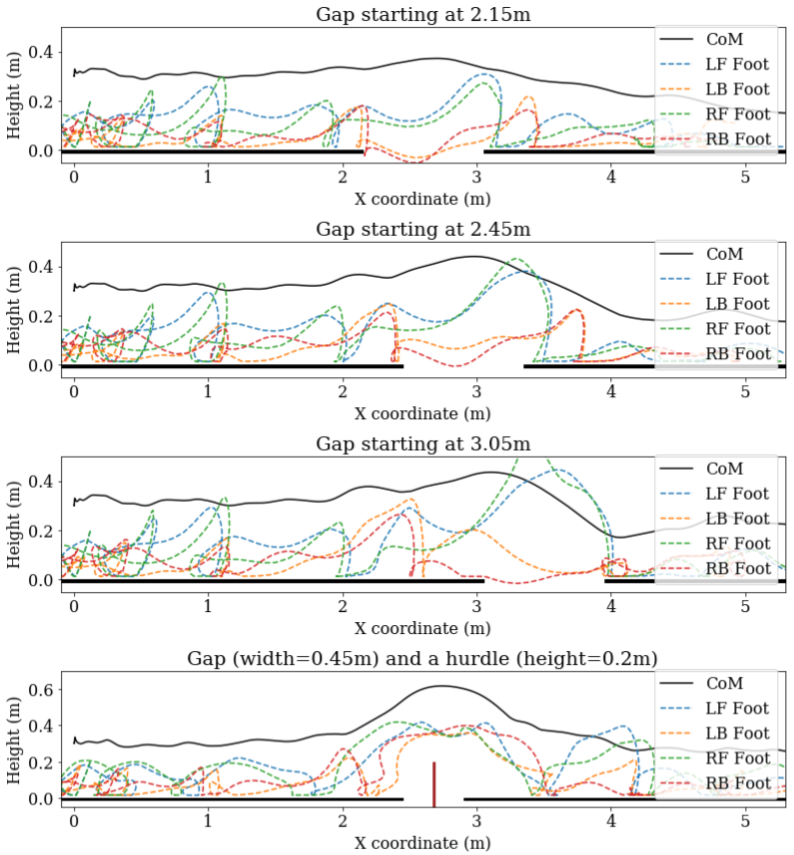}
    \caption{Sample trajectories of the Center of Mass (CoM) and feet of the emerged behaviors. Top 3 plots show the robot jumping over $0.9\si{\metre}$ gaps placed at different locations. The robot adjusts its steps based on the gap location. The last figure shows jumping over a hurdle.}
    \label{fig:com_gap_locs}
\end{figure}

\subsection{Analysis}
The high returns in learning curves are promising, but to deploy the learned gaits on a physical robot, we have to analyze the characteristics of the emerged behavior to determine its feasibility. While we showcase sample behaviors qualitatively in our supplementary video, here we dive deeper into quantitative analysis.

One goal of our generalization is to obtain different agile behaviors by the robot based on different gap or hurdle locations. In Fig.~\ref{fig:com_gap_locs}, we show various behaviors learned by the robot when the gap or hurdle position varies. The plots demonstrate that the robot takes different (lower or higher) jumping trajectories, and adjusts foot placement right before jumping based on the gap or hurdle's location. For example, the robot can cross a gap located at $2.15\si{\metre}$ without extra vertical movement (1st row of Fig. \ref{fig:com_gap_locs}). In contrast, when the gap starts at $3.05\si{\metre}$ (3rd row of Fig. \ref{fig:com_gap_locs}), the robot steps further forward and performs a bigger leap, during which the front feet reach $0.4\si{\metre}$. In the presence of a hurdle (4th row of Fig. \ref{fig:com_gap_locs}), the robot jumps $0.3\si{\metre}$ higher than its normal running height to a peak of $0.6\si{\metre}$.


\begin{figure}[t]
    \centering
    \includegraphics[width=1.0\linewidth]{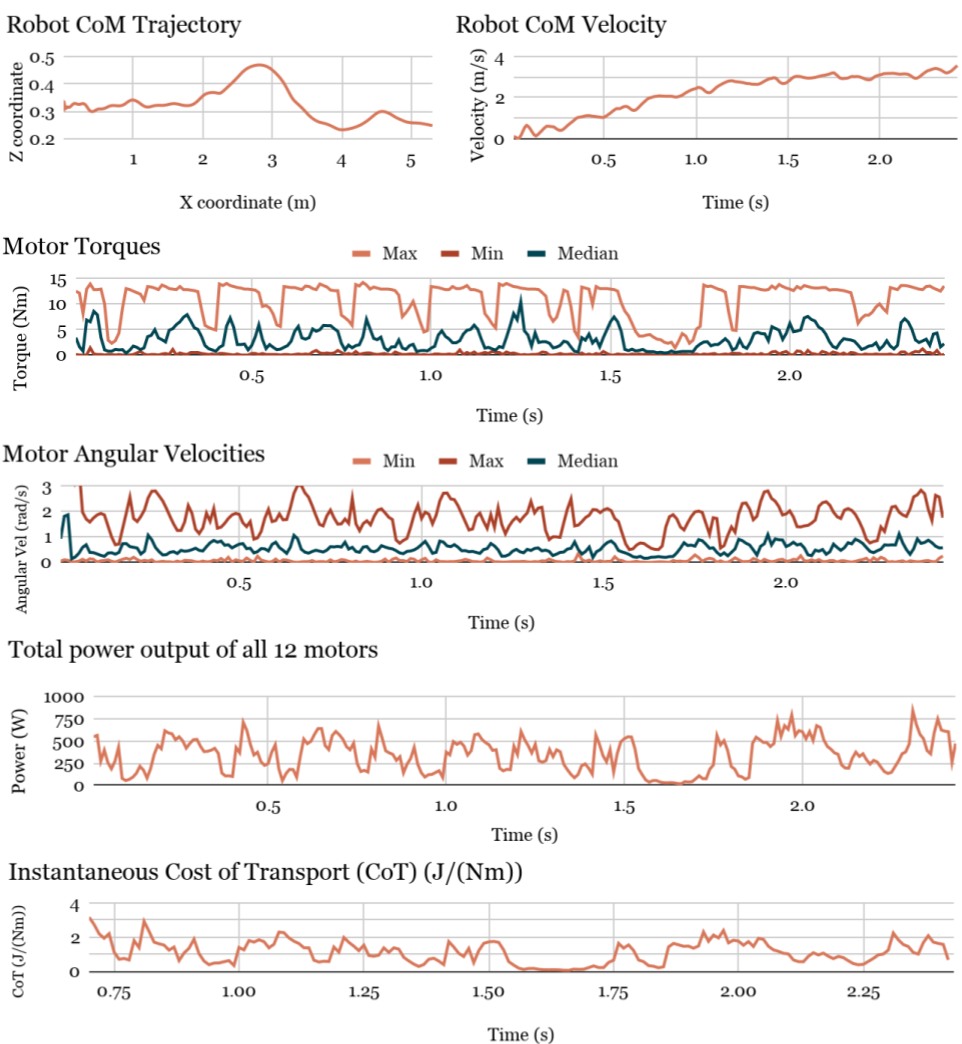}
    \caption{Analysis of one of the evaluations where the robot crosses the gap located at $2.75\si{\metre}$ from the starting point. Flight phase can be seen between $t=1.55\si{\second}$ and $t=1.7\si{\second}$. The Cost of Transport (CoT), calculated as $\frac{\textrm{Power}}{\textrm{Weight*Velocity}}$, lowers with increasing speed. 
    }
    \label{fig:analysis}
\end{figure}


Lastly, we analyze the mechanical characteristics of the emerged behaviors. We take a single episode and look at the motor data, including torque, angular speed and power. Figure \ref{fig:analysis} shows an episode where the gap starts at $2.9\si{\metre}$. The robot performs a big leap to cross the gap, and its CoM is lowered during landing phase. The peak velocity of the robot is roughly $3.2\si{\metre\per\second}$. The average torque of the 12 motors fluctuates between $0$ and $7 \si{\newton\metre}$, and the maximum torque stays below 13 $\si{\newton\metre}$. Cost of Transport ($\frac{\textrm{Power}}{\textrm{Weight} \times \textrm{Velocity}}$) is a common measure used to calculate the energy efficiency of a gait. Once the robot reaches the velocity of $1.0 \si{\metre\per\second}$, the cost of transport during the rest of the episode fluctuates between $0$ and $3\si{\joule\per\newton\per\metre}$. The flight phase of the robot is between $t=1.55\si{\second}$ and $t=1.70\si{\second}$ where both the power and CoT are close to 0. While the learned gaits are optimized for agility rather than energy efficiency, the values that we obtain are reasonable compared to other simulation models such as Cheetah-cub \cite{sprowitz2013towards}. On the other hand, CoT calculated in simulation can be significantly lower than that on robots due to unmodeled gearbox or mechanical inefficiencies \cite{sprowitz2013towards}.

\section{Conclusion}
Legged robots have the potential to perform agile behaviors in order to accomplish complex tasks. In this work, we showed that it is possible for a robot to learn highly agile behaviors using a mentor combined with a three-stage learning process. Our framework consists of finding the right mentor using a simplified task, generalizing to the full problem and finally dropping out the mentor. We tested our framework using PPO and ARS in a realistic physics simulation. Our framework enabled the robot to successfully learn to jump over $0.9\si{\metre}$ gaps and $0.2\si{\metre}$ vertical hurdles. Empirical analysis of the torques, power consumption and speed of the simulated motors showed that the emerged behaviors are efficient and feasible to be deployed on a real robot. Given the successful simulation results, our next step is to transfer these policies to a physical robot with a similar form factor (e.g. Unitree A1). We also plan to extend our repertoire of agile behaviors to tackle parkour on an obstacle course. 




\clearpage

\bibliography{references}  

\begin{thebibliography}{10}
\providecommand{\url}[1]{#1}
\csname url@samestyle\endcsname
\providecommand{\newblock}{\relax}
\providecommand{\bibinfo}[2]{#2}
\providecommand{\BIBentrySTDinterwordspacing}{\spaceskip=0pt\relax}
\providecommand{\BIBentryALTinterwordstretchfactor}{4}
\providecommand{\BIBentryALTinterwordspacing}{\spaceskip=\fontdimen2\font plus
\BIBentryALTinterwordstretchfactor\fontdimen3\font minus
  \fontdimen4\font\relax}
\providecommand{\BIBforeignlanguage}[2]{{%
\expandafter\ifx\csname l@#1\endcsname\relax
\typeout{** WARNING: IEEEtran.bst: No hyphenation pattern has been}%
\typeout{** loaded for the language `#1'. Using the pattern for}%
\typeout{** the default language instead.}%
\else
\language=\csname l@#1\endcsname
\fi
#2}}
\providecommand{\BIBdecl}{\relax}
\BIBdecl

\bibitem{raibert1986legged}
M.~H. Raibert, \emph{Legged robots that balance}, 1986.

\bibitem{park2017high}
H.-W. Park, P.~M. Wensing, and S.~Kim, ``High-speed bounding with the {MIT}
  {Cheetah} 2: Control design and experiments,'' \emph{The International
  Journal of Robotics Research}, vol.~36, no.~2, pp. 167--192, 2017.

\bibitem{hyun2016implementation}
D.~J. Hyun, J.~Lee, S.~Park, and S.~Kim, ``Implementation of trot-to-gallop
  transition and subsequent gallop on the {MIT} {Cheetah} {I},'' \emph{The
  International Journal of Robotics Research}, vol.~35, no.~13, pp. 1627--1650,
  2016.

\bibitem{briggs2012tails}
R.~Briggs, J.~Lee, M.~Haberland, and S.~Kim, ``Tails in biomimetic design:
  {Analysis}, simulation, and experiment,'' in \emph{2012 IEEE/RSJ
  International Conference on Intelligent Robots and Systems}.\hskip 1em plus
  0.5em minus 0.4em\relax IEEE, 2012, pp. 1473--1480.

\bibitem{patel2014rapid}
A.~Patel and M.~Braae, ``Rapid acceleration and braking: {Inspirations} from
  the cheetah's tail,'' in \emph{2014 IEEE International Conference on Robotics
  and Automation (ICRA)}.\hskip 1em plus 0.5em minus 0.4em\relax IEEE, 2014,
  pp. 793--799.

\bibitem{khoramshahi2013piecewise}
M.~Khoramshahi, H.~J. Bidgoly, S.~Shafiee, A.~Asaei, A.~J. Ijspeert, and M.~N.
  Ahmadabadi, ``Piecewise linear spine for speed--energy efficiency trade-off
  in quadruped robots,'' \emph{Robotics and Autonomous Systems}, vol.~61,
  no.~12, pp. 1350--1359, 2013.

\bibitem{eckert2015comparing}
P.~Eckert, A.~Spr{\"o}witz, H.~Witte, and A.~J. Ijspeert, ``Comparing the
  effect of different spine and leg designs for a small bounding quadruped
  robot,'' in \emph{2015 IEEE International Conference on Robotics and
  Automation (ICRA)}.\hskip 1em plus 0.5em minus 0.4em\relax IEEE, 2015, pp.
  3128--3133.

\bibitem{park2015online}
H.-W. Park, P.~M. Wensing, S.~Kim \emph{et~al.}, ``Online planning for
  autonomous running jumps over obstacles in high-speed quadrupeds,'' in
  \emph{Robotics: Science and Systems}, 2015.

\bibitem{nguyen2019optimized}
Q.~{Nguyen}, M.~J. {Powell}, B.~{Katz}, J.~D. {Carlo}, and S.~{Kim},
  ``{Optimized} {Jumping} on the {MIT} {Cheetah} 3 {Robot},'' in \emph{2019
  International Conference on Robotics and Automation (ICRA)}, 2019, pp.
  7448--7454.

\bibitem{bd2018}
``Boston {Dynamics} (2018) {Atlas} - the world’s most dynamic humanoid.''
  \url{https://www.bostondynamics.com/atlas}, 2018.

\bibitem{bledt2020extracting}
G.~{Bledt} and S.~{Kim}, ``{Extracting} {Legged} {Locomotion} {Heuristics} with
  {Regularized} {Predictive} {Control},'' in \emph{2020 IEEE International
  Conference on Robotics and Automation (ICRA)}, 2020, pp. 406--412.

\bibitem{yang2019data}
\BIBentryALTinterwordspacing
Y.~Yang, K.~Caluwaerts, A.~Iscen, T.~Zhang, J.~Tan, and V.~Sindhwani, ``Data
  {Efficient} {Reinforcement} {Learning} for {Legged} {Robots},'' \emph{CoRR},
  vol. abs/1907.03613, 2019. [Online]. Available:
  \url{http://arxiv.org/abs/1907.03613}
\BIBentrySTDinterwordspacing

\bibitem{tan2018sim2real}
\BIBentryALTinterwordspacing
J.~Tan, T.~Zhang, E.~Coumans, A.~Iscen, Y.~Bai, D.~Hafner, S.~Bohez, and
  V.~Vanhoucke, ``Sim-to-real: {Learning} {Agile} {Locomotion} {For}
  {Quadruped} {Robots},'' in \emph{Robotics: Science and Systems}, 2018.
  [Online]. Available: \url{https://arxiv.org/pdf/1804.10332.pdf}
\BIBentrySTDinterwordspacing

\bibitem{hwangbo2019learning}
J.~Hwangbo, J.~Lee, A.~Dosovitskiy, D.~Bellicoso, V.~Tsounis, V.~Koltun, and
  M.~Hutter, ``Learning agile and dynamic motor skills for legged robots,''
  \emph{Science Robotics}, vol.~4, no.~26, 2019.

\bibitem{haarnoja2018learning}
T.~Haarnoja, S.~Ha, A.~Zhou, J.~Tan, G.~Tucker, and S.~Levine, ``Learning to
  walk via deep reinforcement learning,'' in \emph{Robotics: Science and
  Systems}, 2019.

\bibitem{xie2019iterative}
Z.~Xie, P.~Clary, J.~Dao, P.~Morais, J.~Hurst, and M.~van~de Panne, ``Iterative
  reinforcement learning based design of dynamic locomotion skills for
  cassie,'' \emph{arXiv preprint arXiv:1903.09537}, 2019.

\bibitem{RoboImitationPeng20}
X.~B. Peng, E.~Coumans, T.~Zhang, T.-W.~E. Lee, J.~Tan, and S.~Levine,
  ``{Learning} {Agile} {Robotic} {Locomotion} {Skills} by {Imitating}
  {Animals},'' in \emph{Robotics: Science and Systems}, 07 2020.

\bibitem{Lee_2020}
\BIBentryALTinterwordspacing
J.~Lee, J.~Hwangbo, L.~Wellhausen, V.~Koltun, and M.~Hutter, ``Learning
  quadrupedal locomotion over challenging terrain,'' \emph{Science Robotics},
  vol.~5, no.~47, p. eabc5986, Oct 2020. [Online]. Available:
  \url{http://dx.doi.org/10.1126/scirobotics.abc5986}
\BIBentrySTDinterwordspacing

\bibitem{2016-TOG-deepRL}
X.~B. Peng, G.~Berseth, and M.~van~de Panne, ``Terrain-{Adaptive} {Locomotion}
  {Skills} {Using} {Deep} {Reinforcement} {Learning},'' \emph{ACM Transactions
  on Graphics (Proc. SIGGRAPH 2016)}, vol.~35, no.~4, 2016.

\bibitem{peng2015dynamic}
------, ``Dynamic terrain traversal skills using reinforcement learning,''
  \emph{ACM Trans. Graph.}, vol.~34, no.~4, pp. 80:1--80:11, Jul. 2015.

\bibitem{heess2017emergence}
N.~Heess, D.~TB, S.~Sriram, J.~Lemmon, J.~Merel, G.~Wayne, Y.~Tassa, T.~Erez,
  Z.~Wang, S.~Eslami \emph{et~al.}, ``Emergence of locomotion behaviours in
  rich environments,'' \emph{arXiv preprint arXiv:1707.02286}, 2017.

\bibitem{xie2020allsteps}
Z.~Xie, H.~Y. Ling, N.~H. Kim, and M.~van~de Panne, ``{ALLSTEPS}:
  {Curriculum}-driven {Learning} of {Stepping} {Stone} {Skills},'' 2020.

\bibitem{2018-TOG-deepMimic}
\BIBentryALTinterwordspacing
X.~B. Peng, P.~Abbeel, S.~Levine, and M.~van~de Panne, ``{DeepMimic}:
  {Example}-guided {Deep} {Reinforcement} {Learning} of {Physics}-based
  {Character} {Skills},'' \emph{ACM Trans. Graph.}, vol.~37, no.~4, pp.
  143:1--143:14, Jul. 2018. [Online]. Available:
  \url{http://doi.acm.org/10.1145/3197517.3201311}
\BIBentrySTDinterwordspacing

\bibitem{tang2020learning}
Y.~Tang, J.~Tan, and T.~Harada, ``{Learning} {Agile} {Locomotion} via
  {Adversarial} {Training},'' 2020.

\bibitem{chen2019learning}
D.~Chen, B.~Zhou, V.~Koltun, and P.~Krähenbühl, ``Learning by {Cheating},''
  2019.

\bibitem{iscen2018policies}
A.~Iscen, K.~Caluwaerts, J.~Tan, T.~Zhang, E.~Coumans, V.~Sindhwani, and
  V.~Vanhoucke, ``Policies {Modulating} {Trajectory} {Generators},'' in
  \emph{CoRL}, 2018, pp. 916--926.

\bibitem{schulman2017ppo}
J.~Schulman, F.~Wolski, P.~Dhariwal, A.~Radford, and O.~Klimov, ``Proximal
  {Policy} {Optimization} {Algorithms},'' \emph{CoRR}, vol. abs/1707.06347,
  2017.

\bibitem{ars2018}
\BIBentryALTinterwordspacing
H.~Mania, A.~Guy, and B.~Recht, ``Simple random search provides a competitive
  approach to reinforcement learning,'' \emph{CoRR}, vol. abs/1803.07055, 2018.
  [Online]. Available: \url{http://arxiv.org/abs/1803.07055}
\BIBentrySTDinterwordspacing

\bibitem{minicheetah}
B.~{Katz}, J.~D. {Carlo}, and S.~{Kim}, ``Mini {Cheetah}: {A} {Platform} for
  {Pushing} the {Limits} of {Dynamic} {Quadruped} {Control},'' in \emph{2019
  International Conference on Robotics and Automation (ICRA)}, 2019, pp.
  6295--6301.

\bibitem{pybulletcoumans}
E.~Coumans and Y.~Bai, ``Pybullet, a python module for physics simulation for
  games, robotics and machine learning,'' \url{http://pybullet.org},
  2016--2020.

\bibitem{sprowitz2013towards}
A.~Badri-Spröwitz, A.~Tuleu, M.~Vespignani, M.~Ajallooeian, E.~Badri, and
  A.~Ijspeert, ``Towards {Dynamic} {Trot} {Gait} {Locomotion}: {Design},
  {Control}, and {Experiments} with {Cheetah-cub}, a {Compliant} {Quadruped}
  {Robot},'' \emph{The International Journal of Robotics Research}, vol.~32, 07
  2013.

\end{thebibliography}
\bibliographystyle{IEEEtran}

\end{document}